\def\year{2020}\relax
\documentclass[letterpaper]{article} 
\usepackage{aaai20}  
\usepackage{times}  
\usepackage{helvet} 
\usepackage{courier}  
\usepackage[hyphens]{url}  
\usepackage{graphicx} 
\usepackage{graphicx}  
\usepackage{algorithm,algorithmic}
\usepackage{amsmath, amsfonts, amssymb, bm, mathtools, bbm, amsthm}
\title{Crowd Counting with Decomposed Uncertainty}
\author{Min-hwan Oh\textsuperscript{\rm 1}\thanks{Part of the work was done during the internship at IBM.}, Peder Olsen\textsuperscript{\rm 2}\thanks{Much of the work was done while the author was at IBM.}, Karthikeyan Natesan Ramamurthy\textsuperscript{\rm 3}\\ 
\textsuperscript{\rm 1}Columbia University, New York, NY 10025\\
\textsuperscript{\rm 2}Microsoft, Azure Global Research, Redmond, WA 98052\\
\textsuperscript{\rm 3}IBM Research, Yorktown Heights, NY 10598\\
\textsuperscript{\rm 1}m.oh@columbia.edu,  \textsuperscript{\rm 2}peolsen@microsoft.com, \textsuperscript{\rm 3}knatesa@us.ibm.com
}
 
\begin{document}

\maketitle
\begin{abstract}
Research in neural networks in the field of computer vision has achieved remarkable accuracy for point estimation.  However, the uncertainty in the estimation is rarely addressed. Uncertainty quantification accompanied by point estimation can lead to a more informed decision, and even improve the prediction quality. 
In this work, we focus on uncertainty estimation in the domain of crowd counting.  
With increasing occurrences of heavily crowded events such as political rallies, protests, concerts, etc., automated crowd analysis is becoming an increasingly crucial task. The stakes can be very high in many of these real-world applications.
We propose a scalable neural network framework with quantification of decomposed uncertainty using a bootstrap ensemble. We demonstrate that the proposed uncertainty quantification method provides additional insight to the crowd counting problem and is simple to implement. 
We also show that our proposed method exhibits the state of the art performances in many benchmark crowd counting datasets. 
\end{abstract}

\section{Introduction}

The counting problem in computer vision is the estimation of the number of objects in a still image or video frame. It arises in many real-world applications including cell counting in microscopic images \cite{xie2018microscopy}, monitoring crowds in surveillance systems \cite{dollar2012pedestrian}, and counting the number of trees and crops in aerial images \cite{hassaan2016precision,oh2019counting}.
Recently, convolutional neural networks (CNN) have been shown to have successes in a wide range of tasks in computer vision, such as object detection, image recognition, face recognition, and image segmentation. Inspired by these successes, many CNN based crowd counting methods have been proposed. Along with density estimation techniques \cite{lempitsky2010learning}, CNN based approaches have shown outstanding performances over previous works that were relying on handcrafted feature extraction. 

With increasing occurrences of heavily crowded events such as political rallies, protests, concerts, etc., automated crowd analysis is becoming an increasingly crucial task. 
The stakes can be very high in many of these real-world applications.
Hence, we ask the following question:\\

\textit{Can we deploy the existing CNN based crowd counting methods for crowd analysis in real-world applications?}\\

One problem with existing CNN methods is that they only offer point estimates of counts (or density map) and do not address the uncertainty in the prediction, which can come from the model and also from data itself. 
When given a new unlabeled crowd image, how much can we trust the output of the model if it only provides a point estimate?
Probabilistic interpretations of outputs of the predictive model via uncertainty quantification are very important.
Uncertainty quantification accompanied by point estimation can lead to a more informed decision, and even improve the prediction quality.
This can be crucial for the practitioners of these crowd counting methods. With the quantification of prediction confidence at hand, one can treat uncertain inputs and special cases explicitly. For instance, a crowd counting model might return a density map (or a count) with less confidence (i.e., high uncertainty) in some area of a given scene. In this case, the practitioner could decide to pass the image -- or the specific part of the image that the model is uncertain about -- to a human for validation.

While Bayesian methods provide a mathematically plausible framework to deal with uncertainty quantification, often the Bayesian methods for neural networks comes with a prohibitively computational cost. In this work, we propose a simple and scalable neural network framework using a bootstrap ensemble to quantify decomposed uncertainty for crowd counting. The key highlights of our work are:
\begin{itemize}
    \item To the best of our knowledge, this work is the first to address uncertainty quantification of neural network predictions for crowd counting. Our method is shown to produce accurate estimated uncertainty.
    \item Our method decomposes uncertainties coming from the model and also from input data, which can be of independent interest for image analysis.
    \item Our proposed method achieves state-of-the-art level performances on multiple crowd counting benchmark datasets. The proposed architecture is more efficient than previously known state-of-the-art methods in terms of computational complexity.
    \item Our proposed uncertainty quantification framework is generic and independent of the architecture of an underlying network. Combined with its simplicity for implementation, we show the adaptability to other architecture.
\end{itemize}

\section{Related Work}

The previous literature on crowd counting problems can be categorized into three kinds of approaches depending on methodology: detection-based, regression-based and density-based methods.

\textbf{Detection-based crowd counting} is an approach to directly detect each of the target objects in a given image. A typical approach is to utilize object detectors \cite{leibe2005pedestrian,li2008estimating,wang2009crowd} often using moving-windows \cite{dollar2012pedestrian}. Then, the counts of targets in an image are automatically given as a byproduct of detection results. 
However, objects can be highly occluded in many crowded scenes and many target objects can be in drastically different scales, making detection much more challenging. These issues make detection-based approaches infeasible in dense crowd scenes.

\textbf{Regression-based approaches} \cite{chan2009bayesian,chen2012feature,kumagai2017mixture,ryan2009crowd,shang2016end} are proposed to remedy the occlusion problems which are obstacles for detection-based methods.
Regression-based methods directly map input crowd images to scalar values of counts, hence bypassing explicit detection tasks.
Particularly, a mapping between image features and the crowd count is learned. Typically the extracted features are used to generate low-level information, which is learned by a regression model. Hence, these methods leverage better feature extraction (if available) and regression algorithms for estimating counts \cite{shang2016end,arteta2014interactive,chan2009bayesian,chen2012feature,segui2015learning}. 
However, these regression-based methods mostly ignore the spatial information in the crowd images.

\textbf{Density-based crowd counting}, originally proposed in \cite{lempitsky2010learning}, preserves both the count and spatial distribution of the crowd, and have been shown effective at object counting in crowd scenes. In an object density map, the integral over any sub-region is the number of objects within the corresponding region in the image. Density-based methods are generally better at handling cases where objects are severely occluded by bypassing the hard detection of every object, while also maintaining some spatial information about the crowd. \cite{lempitsky2010learning} proposes a method that learns a linear mapping between the image feature and the density map. \cite{pham2015count} proposes learning a non-linear mapping using random forest regression. However, earlier approaches still depended on hand-crafted features.

\textbf{Density-based crowd counting using CNN}.
In recent years, the CNN based methods with density targets have shown performances superior to the traditional methods based on handcrafted features \cite{fu2015fast,wang2015deep,zhang2015cross}. To address perspective issues, \cite{zhang2016single} leverages a multi-column network using convolution filters with different sizes in each column to generate the density map. As a different approach to address perspective issues, \cite{onoro2016towards} proposes taking a pyramid of input patches into a network. \cite{sam2017switching} improves over \cite{zhang2016single} and uses a switching layer to classify the crowd into three classes depending on crowd density and to select one of 3 regressor networks for actual counting.
\cite{zhang2017fcn} incorporates a multi-task objective, jointly estimating the density map and the total count by connecting fully convolutional networks and recurrent networks (LSTM). 
\cite{sindagi2017generating} uses global and local contexts to generate a high-quality density map. 
\cite{li2018csrnet} introduces the dilated convolution to aggregate multi-scale contextual information and utilizes a much deeper architecture from VGG-16 \cite{simonyan2014very}.
\cite{cao2018scale} proposes an encoder-decoder network with the encoder extracting multi-scale features with scale aggregation modules and the decoder generating density maps by using a set of transposed convolutions.

\textbf{Limitations of the current state of the art}: While density estimation and CNN based approaches have shown outstanding performances in the problems of crowd counting, less attention has been paid to assessing uncertainty in predictive outputs. Probabilistic interpretations via uncertainty quantification are important because (1) lack of understanding of model outputs may provide sub-optimal results and (2) neural networks are subject to overfitting, so making decisions based on point prediction alone may provide incorrect predictions with spuriously high confidence.

\section{Uncertainty in Neural Networks}

Much of the previous work on the Bayesian neural network studied uncertainty quantification founded on parametric Bayesian inference \cite{blundell2015weight,gal2016dropout}. In this work, we consider a  non-parametric bootstrap of functions.

\subsection{Bootstrap ensemble}\label{bootstrap}

Bootstrap is a simple technique for producing a distribution over functions with theoretical guarantees \cite{bickel1981some}. It is also general in terms of the class of models that we can accommodate. In its most common form, a bootstrap method takes as input a dataset $\mathcal{D}$ and a function $f_\theta$. We can transform the original dataset $\mathcal{D}$ into $K$ different datasets $\{ \mathcal{D}_k \}_{k=1}^K$'s of cardinality equal to that of the original data $\mathcal{D}$ that is sampled uniformly with replacement.

Then we train $K$ different models. For each model $f_{\theta_k}$, we train the model on the dataset $\mathcal{D}_k$. So each of these models is trained on data from the same distribution but a different dataset. Then if we want to approximate sampling from the distribution of functions, we sample uniformly an integer $k$ from 1 to $K$ and use the corresponding function $f_{\theta_k}$.

In cases of using neural networks as base models $f_{\theta_k}$, bootstrap ensemble maintains a set of $K$ neural networks $\{ f_{\theta_k} \}_{k=1}^K$ independently on $K$ different bootstrapped subsets of the data. It treats each network as independent samples from the weight distribution. In contrast to traditional Bayesian approaches discussed earlier, bootstrapping is a frequentist method, but with the use of the prior distribution, it could approximate the posterior in a simple manner. Also, it scales nicely to high-dimensional spaces, since it only requires point estimates of the weights. However, one major drawback is that computational load increase linearly with respect to the number of base models. In the following section, we discuss how to mitigate this issue and still maintain reasonable uncertainty estimates.

\subsection{Measures of uncertainty}

When we address uncertainty in predictive modeling, there are two major sources of uncertainty \cite{kendall2017uncertainties}:

\begin{enumerate}
    \item \textbf{epistemic uncertainty} is uncertainty due to our lack of knowledge; we are uncertain because we lack understanding. In terms of machine learning, this corresponds to a situation where our model parameters are poorly determined due to a lack of data, so our posterior over parameters is broad.
    \item \textbf{aleatoric uncertainty} is due to genuine stochasticity in the data. In this situation, an uncertain prediction is the best possible prediction. This corresponds to noisy data; no matter how much data the model has seen, if there is inherent noise then the best prediction possible may be a high entropy one.
\end{enumerate}

Note that whether we apply a Bayesian neural network framework or a bootstrap ensemble framework, the kind of uncertainty which is addressed by either of the methods is epistemic uncertainty only. Epistemic uncertainty is often called model uncertainty and it can be explained away given enough data (in theory as data size increases to infinity this uncertainty converges to zero). Addressing aleatoric uncertainty is also crucial for the crowd counting problem since many crowd images do possess inherent noise, occlusions, perspective distortions, etc. that regardless of how much data the model is trained on, there are certain aspects the model is not able to capture. Following \cite{kendall2017uncertainties}, we incorporate both epistemic uncertainty and aleatoric uncertainty in a neural network for crowd counting. We discuss how we operationalize in a scalable manner in the following section.

\subsection{Calibration of Predictive Uncertainty}\label{section:uncertainty_calibration}

Many methods for estimating predictive uncertainty often fail to capture the true distribution of the data \cite{lakshminarayanan2017simple}. For example, a 95\% posterior confidence interval may not contain the true outcome for 95\% of the time. In such a case, the model is considered to be not \textit{calibrated} \cite{kuleshov2018accurate}.
Bootstrap ensemble methods we consider in this work are also not immune to this issue. Hence, we address this by incorporating a technique recently introduced in \cite{kuleshov2018accurate}, which calibrates any regression methods including neural networks. The proposed procedure is inspired by Platt scaling \cite{platt1999probabilistic} which recalibrates the predictions of a pre-trained classifier in a post-processing step.
\cite{kuleshov2018accurate} shows that the recalibration procedure applied to Bayesian models is guaranteed to produce calibrated uncertainty estimates given enough data. 

\section{Proposed Method}

\subsection{Single network with $K$ output heads}

Training and maintaining several independent neural networks is computationally expensive especially when each base network is a large and deep neural network. To remedy this issue, we adopt a single network framework that is scalable for generating bootstrap samples from a large and deep neural network \cite{osband2016deep,oh2019sequential}.
The network consists of a shared architecture --- for example, convolution layers --- with $K$ bootstrapped heads branching off independently. Each head is trained only on its bootstrapped sub-sample of the data as described in Section \ref{bootstrap}. The shared network learns a joint feature representation across all the data, which can provide significant computational advantages at the cost of lower diversity between heads. This type of bootstrap can be trained efficiently in a single forward/backward pass; it can be thought of as a data-dependent dropout, where the dropout mask for each head is fixed for each data point \cite{srivastava2014dropout}.

\subsection{Capturing epistemic uncertainty}
To capture epistemic uncertainty in a neural network, we put a prior distribution over its weights, for example a Gaussian prior: $[\theta_s, \theta_1, ..., \theta_K] \sim \mathcal{N}(0, \tilde{\sigma}^2)$, where $\theta_s$ is the parameter of the shared network and $\theta_1, ..., \theta_K$ are the parameters of bootstrap heads $1,.., K$. Let $x$ be an image input and $y$ be a density output. Without loss of generality, we define our pixel-wise likelihood as a Gaussian with mean given by the model output: $p\left(y|f_{\theta}(x) \right) = \mathcal{N}\left( f_{\theta}(x), \sigma^2 \right)$, with an observation noise variance $\sigma^2$.

For brevity of notations we overload the term $\theta_k = [\theta_k, \theta_s]$ since $\theta_s$ is shared across all samples. For each iteration of training procedure, we sample the model parameter $\Hat{\theta}_k \sim q(\theta)$ where $q(\theta)$ is a bootstrap distribution. In other words, at each iteration we randomly choose which head to use to predict an output $\Hat{y} = f_{\Hat{\theta}_k}(x)$. Then the objective is to minimize the loss (for a single image $x$) given by the negative log-likelihood:
\begin{equation*}\label{eq:epistemicLoss}
    \mathcal{L}(\theta) = \frac{1}{D} \sum_{i} \frac{1}{2\sigma^2} \| y_i -  \Hat{y}_i \|^2 + \frac{1}{2}\log \sigma^2
\end{equation*}
where $y_i$ is the $i$-th pixel of the output density $y$ corresponding to input $x$ and $D$ is the number of output pixels. Note that the observation noise $\sigma^2$ which captures how much noise we have in the outputs stays constant for all data points. Hence we can further drop the second term (since it does not depend on $\theta$), but for the sake of consistency with the following section where we discuss a heteroscedastic setting, we leave it as is. Now, epistemic uncertainty can be captured by the predictive variance, which can be approximated as:
\begin{equation}\label{eq:predictive_var}
    \text{Var}(y) \approx \sigma^2 + \frac{1}{K} \sum_{k=1}^K f_{\Hat{\theta}_k}(x)^\top f_{\Hat{\theta}_k}(x) - \mathbb{E}(y)^\top \mathbb{E}(y)
\end{equation}
with approximated mean: $\mathbb{E}(y) \approx \frac{1}{K}\sum_{k=1}^K f_{\Hat{\theta}_k}(x)$.
Note that during training procedure we randomly select one output head but during test time we combine individual predictions from $K$ heads to compute the predictive mean and the variance.

\subsection{Incorporating aleatoric uncertainty}
In contrast to homoscedastic settings where we assume the observation noise $\sigma^2$ is constant for all inputs, heteroscedastic regression assumes that $\sigma^2$ can vary with input $x$ \cite{le2005heteroscedastic,nix1994estimating}. This change can be useful in cases where parts of the observation space might have higher noise levels than others \cite{kendall2017uncertainties}. In crowd counting applications, it is often the case that images may come from different cameras and scenes. Also due to occlusion and perspective issues within a single image, it is often the case that observation noise can vary from one part of an image (or pixel) to another part (or pixel). 

Following \cite{kendall2017uncertainties}, the network outputs both the estimated density map $y$ and the noise variance $\sigma^2$. Therefore, in our bootstrap implementation of the network, the output layer has a total of $K+1$ nodes --- $K$ nodes corresponding to an ensemble of density map predictions $y$ and an extra node corresponding to $\sigma^2$. Let $\theta_{\sigma}$ be the parameter corresponding to the output node of the noise variance $\sigma^2$. Now, as before, we overload the term $\theta_k = [\theta_k, \theta_s, \theta_\sigma]$ since $\theta_{\sigma}$ is shared across the bootstrap sampling.
We draw a sample of model parameters from the approximate posterior given by bootstrap ensemble $\Hat{\theta}_k \sim q(\theta)$. But this time as described above, we have two parallel outputs, the density map estimate $\Hat{y}$ and the noise variance estimate $\Hat{\sigma}^2$:
\begin{equation*}
[\Hat{y}, \Hat{\sigma}^2] =  f_{\Hat{\theta}_k}(x).
\end{equation*}

Then, we have the following loss given input image $x$ which we want to minimize: 
\begin{equation*}\label{eq:heteroscedastic}
\mathcal{L}(\theta) = \frac{1}{D} \sum_i  \frac{1}{2 \Hat{\sigma}^2_i} \| y_i - \Hat{y}_i \|^2 + \frac{1}{2}\log \Hat{\sigma}^2_i.
\end{equation*}
 Note that this loss contains two parts: the least square residual term which depends on the model uncertainty (epistemic uncertainty) and an aleatoric uncertainty regularization term. Now, if the model predicts $\Hat{\sigma}^2$ to be too high, then the residual term will not have much effect on updating the weights -- the second term will dominate the loss. Hence, the model can learn to ignore the noisy data, but is penalized for that. In practice, due to the numerical stability of predicting $\sigma^2$ which should be positive, we predict the log variance $s_i := \log \sigma^2$ instead of $\sigma^2$ for the output \cite{kendall2017uncertainties}.

\begin{figure*}[t]
    \centering
    \includegraphics[width=\linewidth]{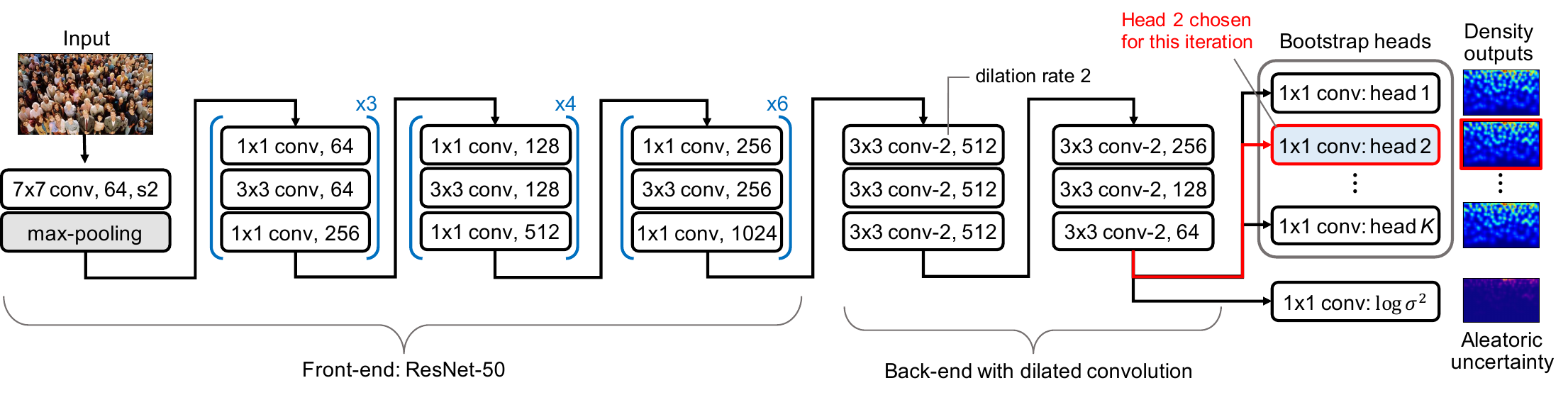}
    \caption{Network architecture of our proposed method, DUBNet. All convolutional layers use stride size 1 and SAME padding to maintain the size of the output the same as the input size. For the front-end, all the convolutional blocks contain identity shortcuts \cite{he2016deep}. Max-pooling layer is applied with 3$\times$3 windows with stride size 2. The back-end layers use dilated kernels with rate 2. The output layer branches out to $K$ bootstrap heads and the extra log-variance output. 
}
    \label{fig:architecture}
\end{figure*}

\subsection{Network architecture}\label{architecure}

The network architecture is composed of three major components: (1) convolutional layers as the front-end for feature extraction, (2) a dilated convolutional for the back-end which uses dilated kernels to deliver larger reception fields and to replace pooling operations, and (3) the bootstrap ensemble output layer.  For the front end, we use ResNet-50 \cite{he2016deep} as backbone. Note that ResNet-50 is much deeper than VGG-16 but ResNet-50 (3.8B FLOPs) has much lower complexity than VGG-16 (15.3B FLOPs).
For discussion on dilated convolution, we refer the readers to  \cite{li2018csrnet}.
For the output layer, we use the $K$ bootstrap ensemble heads for $\hat{y}$ and another output for $\hat{\sigma^2}$. Each output head in the bootstrap ensemble is a fully connected layer. We call our network DUBNet where ``DUB'' stands for decomposed uncertainty using bootstrap ensemble. The details of the architecture are shown in Figure \ref{fig:architecture}.

\subsection{Training procedure}

We initialize the front-end layers (the first 10 convolutional layers) in our model with the corresponding part of a pre-trained ResNet-50 \cite{he2016deep}. For the rest of the parameters, we initialize with a Gaussian distribution with mean 0 and standard deviation 0.01. 
Given a training dataset of input images $X = \{x_1, ..., x_N\}$ and corresponding ground truth density maps $Y = \{y_1, ..., y_N\}$, at each iteration, we sample uniformly at random $k \in \{1, ..., K\}$ to choose an output head $k$ and predict $[\Hat{y}_n, \Hat{s}_n] =  f_{\Hat{\theta}_k}(x_n)$ for $n$-th image as discussed in the previous sections. Algorithm \ref{algo:bootstrap_CNN} presents a single-image batch training procedure. $\Hat{y}_{n,i}$ and $\Hat{s}_{n,i}$ are the $i$-th pixel of the estimated density map and the log variance respectively corresponding to input image $x_n$. $D_n$ is the number of output pixels of $y_n$. Due to pooling operations, the number of output pixels is the same as the number of input pixels.
Adam optimizer \cite{kingma2014adam} with a learning rate of $10^{-5}$ is applied to train the model.

\begin{algorithm}
\caption{Decomposed Uncertainty using Bootstrap}
\begin{algorithmic}[1]
    \REQUIRE Input images $\{x_n\}_{n=1}^N$, GT density $\{y_n\}_{n=1}^N$
    \STATE Initialize parameters $\theta$
    \FOR{each epoch}
        \FOR{all $n = 1$ to $N$}
            \STATE Sample a bootstrap head $k \sim \text{Uniform}\{1,...,K\}$
            \STATE Compute predictions $[\Hat{y}_n, \Hat{s}_n] =  f_{\Hat{\theta}_k}(x_n)$
            \STATE Compute loss:
            \vspace{-0.5em}
            \begin{equation*}\vspace{-0.5em}
                \resizebox{.85\hsize}{!}{ $ \mathcal{L}(\theta_k) = \frac{1}{D_n} \sum_i  \frac{1}{2 \exp(\Hat{s}_{n,i})} \| y_{n,i} - \Hat{y}_{n,i} \|^2 + \frac{1}{2} \Hat{s}_{n,i} $}
                \end{equation*}
            \STATE Update $\theta_k$ using gradient $\frac{d\mathcal{L}(\theta_k)}{d\theta_k}$
        \ENDFOR
    \ENDFOR
\end{algorithmic}
\label{algo:bootstrap_CNN}
\end{algorithm}

\subsection{Recalibration of Predictive Uncertainty}\label{section:recalibration_details}

Once we have a trained model $f_{\hat{\theta}}$, we compute the mean prediction $\mu(x_n) = \frac{1}{K} \sum_k f_{\hat{\theta}_k}(x_n)$ for an input image $x_n$. Note that $\mu(x_n)$ is a density map. We sum over all pixels in $\mu(x_n)$ to compute the predicted mean count $\bar{C}_n$. Similarly, using the (pixel-wise) predictive variance in Eq.\eqref{eq:predictive_var}, we compute the predictive standard deviation in counts $\bar{\sigma}_n$ by summing over all pixels. Then we construct a standardized residual $Z_n = (C_n - \bar{C}_n)/\bar{\sigma}_n$ where $C_n$ is the ground-truth count for image $x_n$ and construct a quantile target $\hat{P}(Z_n)$ which is the proportion of data whose standardized residual is below $Z_n$. Then using each pair $(Z_n , \hat{P}(Z_n) )$, we fit an isotonic regression model $\mathcal{R}$. The recalibration procedure is summarized in Algorithm \ref{algo:recalibration}.

\begin{algorithm}
\caption{Uncertainty Recalibration}
\begin{algorithmic}[1]
    \REQUIRE $\{C_n, \bar{C}_n, \bar{\sigma}\}_{n=1}^N$ for validation data
    \STATE Compute $Z_n = (C_n - \bar{C}_n)/\bar{\sigma}_n$ for all $n$
    \STATE Construct a recalibration dataset:
    \vspace{-0.5em}
    \begin{equation*}\vspace{-0.8em}
        \tilde{\mathcal{D}} = \left\{ \left( Z_n , \hat{P}(Z_n) \right)  \right\}_{n=1,..., N}
    \end{equation*}
    where
    $\hat{P}(z) = |\{ C_m \mid Z_m \leq z, m = 1, ..., N \}| / N$
    \STATE Train a isotonic regression model $\mathcal{R}$ on $\tilde{\mathcal{D}} $.
\end{algorithmic}
\label{algo:recalibration}
\end{algorithm}

Note that the recalibration dataset $\tilde{\mathcal{D}}$ is constructed using a validation data (non-training data) and the model $\mathcal{R}$
is fitted on this dataset. Once $\mathcal{R}$ is learned, for a given quantile $p$, (e.g. 0.95 and 0.05 for 90\% confidence interval) one can easily find $Z^p \in \mathbb{R}$ such that $\mathcal{R}(Z^p) \approx p$ (since $\mathcal{R}$ is a monotone function).
When using at a test time where we only have $\bar{C}_n$ and $\bar{\sigma}_n$, we can construct a confidence bound by computing $\bar{C}_n + \bar{\sigma}_n Z^p$.

\section{Experiments}
In this section, we first introduce datasets and experiment details. We give the evaluation results and perform comparisons between the proposed method with recent state-of-the-art methods. For all experiments, we used $K=10$ heads for DUBNet. We follow the standard procedure to generate the ground truth density map using Gaussian kernels.

\subsection{Evaluation metrics}
For crowd counting evaluation, the count estimation error is measured by two metrics, Mean Absolute Error (MAE) and Root Mean Squared Error (RMSE), which are commonly used for quantitative comparison in previous works. They are defined as follows:
\begin{align*}
  \resizebox{.97\hsize}{!}{ $\displaystyle \mathrm{MAE} = \frac{1}{N}\sum_{n=1}^N | \Hat{C}_n - C_n |, \quad \mathrm{RMSE} = \sqrt{ \frac{1}{N}\sum_{n=1}^N ( \Hat{C}_n - C_n )^2 } $}
\end{align*}
where $N$ is the number of test samples, $C_n$ is the true crowd count for the $n$-th image sample and $\Hat{C}_n$ is the corresponding estimated count. $C_n$ and $\Hat{C}_n$ are given by the
integration over the ground truth density map $\sum_i y_{n,i}$ and over an estimated density map $\sum_i \Hat{y}_{n,i}$ respectively, where $i$ is the $i$-th pixel in output images. Note that during test time we use predictive mean over $K$ bootstrap outputs as $\hat{y}$.

\subsection{Ablation study}
We performed ablation studies on UCF-CC 50 and UCF-QNRF datasets to validate the efficacy of our proposed method. We first compare our proposed architecture DUBNet with its variant which does not have a bootstrap ensemble output and an aleatoric uncertainty output; hence has a single fully connected layer in the output layer. We call this variant "DUBNet w/o DUB extension."

\begin{table}[H]
\begin{center}
\caption{Ablation studies on bootstrap extension and adaptability on different architecture}
\begin{tabular}{|l|c|c|}
\hline
& UCF-50 & UCF-QNRF\\
Methods & MAE & MAE \\
\hline
DUBNet w/o \textbf{DUB} extension & 258.4 & 120.9\\
DUBNet & \textbf{243.8} & \textbf{105.6}\\
\hline
CSRNet (Li et al 2018)  & 266.1 & 135.5 \\
CSRNet + \textbf{DUB} extension & \textbf{244.9}  & \textbf{127.1}  \\
\hline
MCNN \cite{zhang2016single} & 377.6  & 277.0  \\
MCNN + \textbf{DUB} extension &  \textbf{359.4} & \textbf{254.6}  \\
\hline
\end{tabular}
\end{center}
\label{ablation}
\end{table}

To test the adaptability of our framework to other architecture, we applied the DUB extension to CSRNet \cite{li2018csrnet} with branching outputs and an aleatoric uncertainty output. We also applied the same extension to MCNN \cite{zhang2016single}. We compare with the vanilla CSRNet and MCNN respectively.
The results in Table \ref{ablation} show that our proposed framework combining both aleatoric and epistemic uncertainty contributes significantly to performance on the evaluation data, and can be applied to other architecture.

\begin{table*}[t]
\begin{center}
\caption{Estimation errors on ShanghaiTech A/B, UCF-CC 50, and UCF-QNRF datasets}
\begin{tabular}{|l|c|c|c|c|c|c|c|c|}
\hline
&  \multicolumn{2}{c|}{ShanghaiTech A} &  \multicolumn{2}{c|}{ShanghaiTech B} & \multicolumn{2}{c|}{UCF-CC 50} & \multicolumn{2}{c|}{UCF-QNRF}\\
Method & \small MAE & \small RMSE &  \small MAE & \small RMSE & \small MAE & \small RMSE & \small MAE & \small RMSE\\
\hline
MCNN \cite{zhang2016single} & 110.2 & 173.2 & 26.4 & 41.3 & 377.6 & 509.1 & 277 & 426\\
Cascaded-MTL \cite{sindagi2017cnn} & 101.3 & 152.4 & 20.0 & 31.1 & 322.8 & 397.9 & 252 & 514\\
Switch-CNN \cite{sam2017switching} & 90.4 & 135.0 & 21.6 & 33.4 & 318.1 & 439.2 & 228 & 445\\
D-ConvNet \cite{shi2018crowd} & 73.5 & 112.3 & 18.7 & 26.0 & 288.4 & 404.7 & - & -\\
L2R \cite{liu2018leveraging}  & 73.6 & 112.0 & 13.7 & 21.4 & 279.6 & 388.9 & - & -\\
DensityNetwork \cite{idrees2018composition} & - & - & - & - & - & - & 132 & 191\\
CSRNet \cite{li2018csrnet} & 68.2 & 115.0 & 10.6 & 16.0 & 266.1 & 397.5 & 135.5 & 207.4\\
ic-CNN \cite{ranjan2018iterative} & 68.5 & 116.2 & 10.7    & 16.0  & 260.9 & 365.5 & - & -\\
SANet \cite{cao2018scale} & 67.0 & \textbf{104.5} & 8.4 & 13.6 & 258.4 & 334.9 & - & -\\
SFCN \cite{wang2019learning} & 64.8 & 107.5 & \textbf{7.6} & 13.0 & \textbf{214.2} & \textbf{318.2} & \textbf{102.0} & \textbf{171.4}\\
CAN \cite{liu2019context} & \textbf{62.3} & \textbf{100.0} & 7.8 & \textbf{12.2} & \textbf{212.2} & \textbf{243.7} & 107 & 183\\
DUBNet {\small (Ours)} & \textbf{64.6} & 106.8 & \textbf{7.7} & \textbf{12.5} & 243.8 & 329.3 & \textbf{105.6} & \textbf{180.5}\\
\hline
\end{tabular}
\end{center}
\label{Shanghai_table}
\end{table*}

\subsection{Performance comparisons}
We evaluate our method on four publicly available crowd counting datasets: ShanghaiTech \cite{zhang2016single}, UCF-CC 50 \cite{idrees2015detecting}, and UCF-QNRF \cite{idrees2018composition}.
For all datasets, we generate ground truth density maps with fixed spread Gaussian kernel. We compare our method with previously published work. In each table, the previous work which provided code or have been validated by a third party other than the original authors have been listed above our method. For completeness, we also list the recent work (without code or validation by a third party) below our method and include the numbers reported by the original authors. We highlight the best two performances in each metric.

\begin{figure*}[t]
    \centering
    \includegraphics[width=0.99\linewidth]{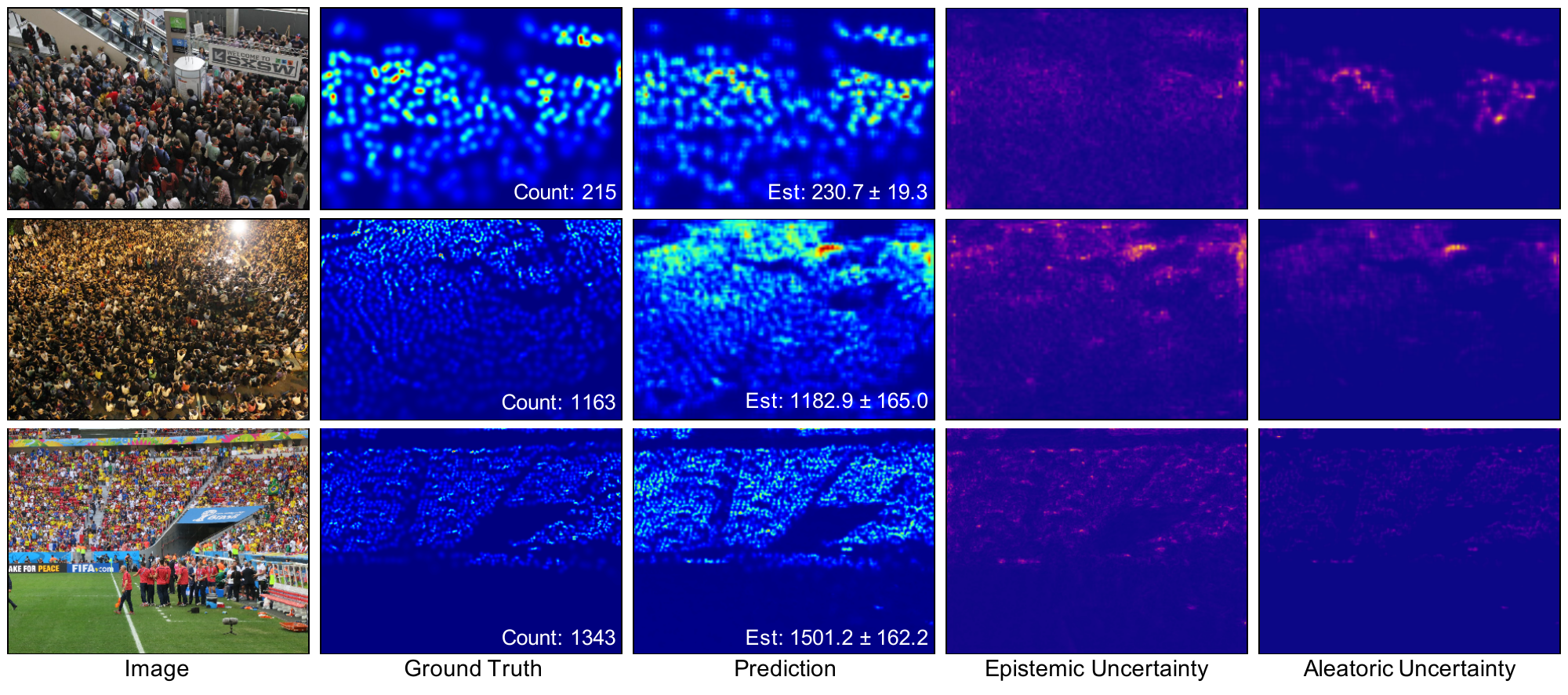}
    \caption{Qualitative results of DUBNet on the ShanghaiTech and the UCF-QNRF datasets. For each image, we demonstrate the ground truth density maps and counts, the estimated density maps and estimated counts with 90\% confidence interval. We also present both estimated epistemic and aleatoric uncertainty quantification. More red color means higher uncertainty. Epistemic uncertainty captures the model's lack of knowledge about the data. Aleatoric uncertainty captures inherent noise in the data.
    }
     \label{fig:Shanghai_A}
\end{figure*}

\textbf{ShanghaiTech}. The ShanghaiTech dataset \cite{zhang2016single} contains 1198 annotated images with a total of 330,165 persons. This dataset consists of two parts: Part A which contains 482 images and Part B which contains 716 images. Part A is randomly collected from the Internet and contains mostly highly congested scenes. Part B contains images captured from street views with relatively sparse crowd scenes. We use the training and testing splits provided by the authors: 300 images for training and 182 images for testing in Part A; 400 images for training and 316 images for testing in Part B.

\textbf{UCF-CC 50}. The UCF-CC 50 dataset \cite{zhang2015cross} is a small dataset which contains only 50 annotated crowd images. However, the challenging aspect of this dataset is that there is a large variation in crowd counts which range from 94 to 4543. Along with this variation, the limited number of images makes it a challenging dataset for the crowd counting tasks. Since training and test data split is not provided, as done in the previous literature \cite{li2018csrnet,zhang2015cross}, We use 5-fold cross-validation to evaluate the performance of the proposed method.

\textbf{UCF-QNRF}. The UCF-QNRF dataset was recently introduced by \cite{idrees2018composition}. It is currently the largest crowd dataset which contains 1,535 images with dense crowds with many of them being high-resolution images. Approximately 1.25 million people were annotated with dot annotations These images come with a wider variety of scenes and contains the most diverse set of viewpoints, densities, and lighting variations. The ground truth counts of the images in the dataset range from 49 to 12,865. Meanwhile, the median and the mean counts are 425 and 815.4, respectively. 
The training dataset contains 1,201 images, with which we train our model. Some of the images are so high-resolution that we faced memory issues in GPU while training. Hence, we down-sampled images that contain more than 3 million pixels. Then, we test our model on the remaining 334 images in the test dataset.

\subsubsection{Results}
The results in Table \ref{Shanghai_table} show that our proposed method is within the top two performers (highlighted in bold) for almost all the benchmark datasets we consider. One thing to note is that our uncertainty quantification framework is not limited to the proposed architecture, but is potentially adaptable to the other state of art architecture (as shown in the ablation study).

\subsection{Estimated uncertainty validation}
Estimated uncertainty is meaningful if it can capture the true distribution of the data. As mentioned earlier in the paper, we can validate whether the estimated uncertainty is well-calibrated or not by checking whether the estimated $p$ quantile confidence interval (CI) contains the true outcome $p$ fraction of the time. Table \ref{CI} shows the fraction of test data in each dataset whose ground truth falls in 90\% CI.\footnote{UCF-CC 50 dataset is not included since the uncertainty recalibration is difficult to perform due to the limited data size.} The results suggest that our estimated uncertainty is accurate.

\begin{table}[ht]
\begin{center}
\caption{Calibration results of estimated uncertainty}
\begin{tabular}{|l|c|}
\hline
Dataset & Ground truth in 90\% CI \\
\hline
ShanghaiTech A &  0.907 \\
ShanghaiTech B &  0.915 \\
UCF-QNRF &  0.890 \\
\hline
\end{tabular}
\end{center}
\label{CI}
\end{table}

\subsection{Discussion on estimated uncertainty}
Figure \ref{fig:Shanghai_A}  visualize the samples along with estimated density maps and their epistemic and aleatoric uncertainty from test evaluations on the ShanghaiTech data and the UCF-QNRF data.
The results demonstrate that the model is generally less confident (i.e. higher epistemic uncertainty) in dense crowd regions of the images, which is natural. There appears to be a certain level of a positive correlation between epistemic and aleatoric uncertainty which is expected -- since the common issues in crowd images such as occlusion and perspective issues are typically correlated with higher crowd density, this can cause both epistemic and aleatoric uncertainty to be higher. But, we also observe a notable difference in the estimated measures of uncertainty in the samples. We observe that aleatoric uncertainty is more prominent in areas where the image itself has more noise (for example, lighting glare in the second image in Figure \ref{fig:Shanghai_A}) and occlusions (right side along the horizontal centerline in the first image in Figure \ref{fig:Shanghai_A}). We can observe that even in very crowded scenes, when occlusions and noise are less prominent, the estimated aleatoric uncertainty can be low -- for example, the stadium image (the third image in Figure \ref{fig:Shanghai_A}) shows very low aleatoric uncertainty over the entire image since there are rarely occlusions or perspective issues due to the stadium seating configuration.

\section{Conclusion}
In this paper, we present a scalable and effective framework that can incorporate uncertainty quantification in prediction for crowd counting. 
The main component of the framework is combining shared convolutional layers and bootstrap ensembles to quantify uncertainty which is decomposed into epistemic and aleatoric uncertainty.
Our proposed framework is generic, independent of the architecture choices, and also easily adaptable to other CNN based crowd counting methods.
The extensive experiments demonstrate that the proposed method, DUBNet, has the state-of-the-art level performance on all benchmark datasets considered, and produces calibrated and meaningful uncertainty estimates.

{\small
\bibliographystyle{aaai}
\bibliography{egbib}

\begin{thebibliography}{}

\bibitem[\protect\citeauthoryear{Arteta \bgroup et al\mbox.\egroup
  }{2014}]{arteta2014interactive}
Arteta, C.; Lempitsky, V.; Noble, J.~A.; and Zisserman, A.
\newblock 2014.
\newblock Interactive object counting.
\newblock In {\em European Conference on Computer Vision},  504--518.
\newblock Springer.

\bibitem[\protect\citeauthoryear{Bickel and Freedman}{1981}]{bickel1981some}
Bickel, P.~J., and Freedman, D.~A.
\newblock 1981.
\newblock Some asymptotic theory for the bootstrap.
\newblock {\em The Annals of Statistics}  1196--1217.

\bibitem[\protect\citeauthoryear{Blundell \bgroup et al\mbox.\egroup
  }{2015}]{blundell2015weight}
Blundell, C.; Cornebise, J.; Kavukcuoglu, K.; and Wierstra, D.
\newblock 2015.
\newblock Weight uncertainty in neural networks.
\newblock {\em arXiv preprint arXiv:1505.05424}.

\bibitem[\protect\citeauthoryear{Cao \bgroup et al\mbox.\egroup
  }{2018}]{cao2018scale}
Cao, X.; Wang, Z.; Zhao, Y.; and Su, F.
\newblock 2018.
\newblock Scale aggregation network for accurate and efficient crowd counting.
\newblock In {\em Proceedings of the European Conference on Computer Vision
  (ECCV)},  734--750.

\bibitem[\protect\citeauthoryear{Chan and Vasconcelos}{2009}]{chan2009bayesian}
Chan, A.~B., and Vasconcelos, N.
\newblock 2009.
\newblock Bayesian poisson regression for crowd counting.
\newblock In {\em Computer Vision, 2009 IEEE 12th International Conference on},
   545--551.
\newblock IEEE.

\bibitem[\protect\citeauthoryear{Chen \bgroup et al\mbox.\egroup
  }{2012}]{chen2012feature}
Chen, K.; Loy, C.~C.; Gong, S.; and Xiang, T.
\newblock 2012.
\newblock Feature mining for localised crowd counting.
\newblock In {\em BMVC}, volume~1, ~3.

\bibitem[\protect\citeauthoryear{CNN.com}{}]{CNNgigapixel}
CNN.com.
\newblock {Gigapixel: the Inauguration of Donald Trump}.
\newblock
  https://edition.cnn.com/interactive/2017/01/politics/trump-inauguration-gigapixel/.
\newblock Accessed: 2018-11-15.

\bibitem[\protect\citeauthoryear{Dollar \bgroup et al\mbox.\egroup
  }{2012}]{dollar2012pedestrian}
Dollar, P.; Wojek, C.; Schiele, B.; and Perona, P.
\newblock 2012.
\newblock Pedestrian detection: An evaluation of the state of the art.
\newblock {\em IEEE transactions on pattern analysis and machine intelligence}
  34(4):743--761.

\bibitem[\protect\citeauthoryear{Fu \bgroup et al\mbox.\egroup
  }{2015}]{fu2015fast}
Fu, M.; Xu, P.; Li, X.; Liu, Q.; Ye, M.; and Zhu, C.
\newblock 2015.
\newblock Fast crowd density estimation with convolutional neural networks.
\newblock {\em Engineering Applications of Artificial Intelligence} 43:81--88.

\bibitem[\protect\citeauthoryear{Gal and Ghahramani}{2016}]{gal2016dropout}
Gal, Y., and Ghahramani, Z.
\newblock 2016.
\newblock Dropout as a bayesian approximation: Representing model uncertainty
  in deep learning.
\newblock In {\em international conference on machine learning},  1050--1059.

\bibitem[\protect\citeauthoryear{Graves}{2011}]{graves2011practical}
Graves, A.
\newblock 2011.
\newblock Practical variational inference for neural networks.
\newblock In {\em Advances in neural information processing systems},
  2348--2356.

\bibitem[\protect\citeauthoryear{Hassaan \bgroup et al\mbox.\egroup
  }{2016}]{hassaan2016precision}
Hassaan, O.; Nasir, A.~K.; Roth, H.; and Khan, M.~F.
\newblock 2016.
\newblock Precision forestry: trees counting in urban areas using visible
  imagery based on an unmanned aerial vehicle.
\newblock {\em IFAC-PapersOnLine} 49(16):16--21.

\bibitem[\protect\citeauthoryear{He \bgroup et al\mbox.\egroup
  }{2016}]{he2016deep}
He, K.; Zhang, X.; Ren, S.; and Sun, J.
\newblock 2016.
\newblock Deep residual learning for image recognition.
\newblock In {\em Proceedings of the IEEE conference on computer vision and
  pattern recognition},  770--778.

\bibitem[\protect\citeauthoryear{Idrees \bgroup et al\mbox.\egroup
  }{2018}]{idrees2018composition}
Idrees, H.; Tayyab, M.; Athrey, K.; Zhang, D.; Al-Maadeed, S.; Rajpoot, N.; and
  Shah, M.
\newblock 2018.
\newblock Composition loss for counting, density map estimation and
  localization in dense crowds.
\newblock {\em arXiv preprint arXiv:1808.01050}.

\bibitem[\protect\citeauthoryear{Idrees, Soomro, and
  Shah}{2015}]{idrees2015detecting}
Idrees, H.; Soomro, K.; and Shah, M.
\newblock 2015.
\newblock Detecting humans in dense crowds using locally-consistent scale prior
  and global occlusion reasoning.
\newblock {\em IEEE transactions on pattern analysis and machine intelligence}
  37(10):1986--1998.

\bibitem[\protect\citeauthoryear{Kendall and
  Gal}{2017}]{kendall2017uncertainties}
Kendall, A., and Gal, Y.
\newblock 2017.
\newblock What uncertainties do we need in bayesian deep learning for computer
  vision?
\newblock In {\em Advances in neural information processing systems},
  5574--5584.

\bibitem[\protect\citeauthoryear{Kingma and Ba}{2014}]{kingma2014adam}
Kingma, D.~P., and Ba, J.
\newblock 2014.
\newblock Adam: A method for stochastic optimization.
\newblock {\em arXiv preprint arXiv:1412.6980}.

\bibitem[\protect\citeauthoryear{Kingma, Salimans, and
  Welling}{2015}]{kingma2015variational}
Kingma, D.~P.; Salimans, T.; and Welling, M.
\newblock 2015.
\newblock Variational dropout and the local reparameterization trick.
\newblock In {\em Advances in Neural Information Processing Systems},
  2575--2583.

\bibitem[\protect\citeauthoryear{Kuleshov, Fenner, and
  Ermon}{2018}]{kuleshov2018accurate}
Kuleshov, V.; Fenner, N.; and Ermon, S.
\newblock 2018.
\newblock Accurate uncertainties for deep learning using calibrated regression.
\newblock In {\em International Conference on Machine Learning},  2801--2809.

\bibitem[\protect\citeauthoryear{Kumagai, Hotta, and
  Kurita}{2017}]{kumagai2017mixture}
Kumagai, S.; Hotta, K.; and Kurita, T.
\newblock 2017.
\newblock Mixture of counting {CNN}s: Adaptive integration of {CNN}s
  specialized to specific appearance for crowd counting.
\newblock {\em arXiv preprint arXiv:1703.09393}.

\bibitem[\protect\citeauthoryear{Lakshminarayanan, Pritzel, and
  Blundell}{2017}]{lakshminarayanan2017simple}
Lakshminarayanan, B.; Pritzel, A.; and Blundell, C.
\newblock 2017.
\newblock Simple and scalable predictive uncertainty estimation using deep
  ensembles.
\newblock In {\em Advances in Neural Information Processing Systems},
  6402--6413.

\bibitem[\protect\citeauthoryear{Le, Smola, and
  Canu}{2005}]{le2005heteroscedastic}
Le, Q.~V.; Smola, A.~J.; and Canu, S.
\newblock 2005.
\newblock Heteroscedastic gaussian process regression.
\newblock In {\em Proceedings of the 22nd international conference on Machine
  learning},  489--496.
\newblock ACM.

\bibitem[\protect\citeauthoryear{Leibe, Seemann, and
  Schiele}{2005}]{leibe2005pedestrian}
Leibe, B.; Seemann, E.; and Schiele, B.
\newblock 2005.
\newblock Pedestrian detection in crowded scenes.
\newblock In {\em null},  878--885.
\newblock IEEE.

\bibitem[\protect\citeauthoryear{Lempitsky and
  Zisserman}{2010}]{lempitsky2010learning}
Lempitsky, V., and Zisserman, A.
\newblock 2010.
\newblock Learning to count objects in images.
\newblock In {\em Advances in neural information processing systems},
  1324--1332.

\bibitem[\protect\citeauthoryear{Li \bgroup et al\mbox.\egroup
  }{2008}]{li2008estimating}
Li, M.; Zhang, Z.; Huang, K.; and Tan, T.
\newblock 2008.
\newblock Estimating the number of people in crowded scenes by mid based
  foreground segmentation and head-shoulder detection.
\newblock In {\em Pattern Recognition, 2008. ICPR 2008. 19th International
  Conference on},  1--4.
\newblock IEEE.

\bibitem[\protect\citeauthoryear{Li, Zhang, and Chen}{2018}]{li2018csrnet}
Li, Y.; Zhang, X.; and Chen, D.
\newblock 2018.
\newblock Csrnet: Dilated convolutional neural networks for understanding the
  highly congested scenes.
\newblock In {\em Computer Vision and Pattern Recognition (CVPR)}.

\bibitem[\protect\citeauthoryear{Liu, Salzmann, and Fua}{2019}]{liu2019context}
Liu, W.; Salzmann, M.; and Fua, P.
\newblock 2019.
\newblock Context-aware crowd counting.
\newblock In {\em Proceedings of the IEEE Conference on Computer Vision and
  Pattern Recognition},  5099--5108.

\bibitem[\protect\citeauthoryear{Liu, van~de Weijer, and
  Bagdanov}{2018}]{liu2018leveraging}
Liu, X.; van~de Weijer, J.; and Bagdanov, A.~D.
\newblock 2018.
\newblock Leveraging unlabeled data for crowd counting by learning to rank.
\newblock In {\em Proceedings of the IEEE Conference on Computer Vision and
  Pattern Recognition},  7661--7669.

\bibitem[\protect\citeauthoryear{Louizos and
  Welling}{2016}]{louizos2016structured}
Louizos, C., and Welling, M.
\newblock 2016.
\newblock Structured and efficient variational deep learning with matrix
  gaussian posteriors.
\newblock In {\em International Conference on Machine Learning},  1708--1716.

\bibitem[\protect\citeauthoryear{Louizos and
  Welling}{2017}]{louizos2017multiplicative}
Louizos, C., and Welling, M.
\newblock 2017.
\newblock Multiplicative normalizing flows for variational bayesian neural
  networks.
\newblock {\em arXiv preprint arXiv:1703.01961}.

\bibitem[\protect\citeauthoryear{MacKay}{1992}]{mackay1992practical}
MacKay, D.~J.
\newblock 1992.
\newblock A practical bayesian framework for backpropagation networks.
\newblock {\em Neural computation} 4(3):448--472.

\bibitem[\protect\citeauthoryear{Maeda}{2014}]{maeda2014bayesian}
Maeda, S.-i.
\newblock 2014.
\newblock A bayesian encourages dropout.
\newblock {\em arXiv preprint arXiv:1412.7003}.

\bibitem[\protect\citeauthoryear{Neal}{1993}]{neal1993bayesian}
Neal, R.~M.
\newblock 1993.
\newblock Bayesian learning via stochastic dynamics.
\newblock In {\em Advances in neural information processing systems},
  475--482.

\bibitem[\protect\citeauthoryear{Nix and Weigend}{1994}]{nix1994estimating}
Nix, D.~A., and Weigend, A.~S.
\newblock 1994.
\newblock Estimating the mean and variance of the target probability
  distribution.
\newblock In {\em Neural Networks, 1994. IEEE World Congress on Computational
  Intelligence., 1994 IEEE International Conference On}, volume~1,  55--60.
\newblock IEEE.

\bibitem[\protect\citeauthoryear{Oh and Iyengar}{2019}]{oh2019sequential}
Oh, M.-h., and Iyengar, G.
\newblock 2019.
\newblock Sequential anomaly detection using inverse reinforcement learning.
\newblock In {\em Proceedings of the 25th ACM SIGKDD International Conference
  on Knowledge Discovery \& Data Mining},  1480--1490.

\bibitem[\protect\citeauthoryear{Oh, Olsen, and
  Ramamurthy}{2019}]{oh2019counting}
Oh, M.-h.; Olsen, P.; and Ramamurthy, K.~N.
\newblock 2019.
\newblock Counting and segmenting sorghum heads.
\newblock {\em arXiv preprint arXiv:1905.13291}.

\bibitem[\protect\citeauthoryear{Onoro-Rubio and
  L{\'o}pez-Sastre}{2016}]{onoro2016towards}
Onoro-Rubio, D., and L{\'o}pez-Sastre, R.~J.
\newblock 2016.
\newblock Towards perspective-free object counting with deep learning.
\newblock In {\em European Conference on Computer Vision},  615--629.
\newblock Springer.

\bibitem[\protect\citeauthoryear{Osband \bgroup et al\mbox.\egroup
  }{2016}]{osband2016deep}
Osband, I.; Blundell, C.; Pritzel, A.; and Van~Roy, B.
\newblock 2016.
\newblock Deep exploration via bootstrapped dqn.
\newblock In {\em Advances in neural information processing systems},
  4026--4034.

\bibitem[\protect\citeauthoryear{Pham \bgroup et al\mbox.\egroup
  }{2015}]{pham2015count}
Pham, V.-Q.; Kozakaya, T.; Yamaguchi, O.; and Okada, R.
\newblock 2015.
\newblock Count forest: Co-voting uncertain number of targets using random
  forest for crowd density estimation.
\newblock In {\em Proceedings of the IEEE International Conference on Computer
  Vision},  3253--3261.

\bibitem[\protect\citeauthoryear{Platt and
  others}{1999}]{platt1999probabilistic}
Platt, J., et~al.
\newblock 1999.
\newblock Probabilistic outputs for support vector machines and comparisons to
  regularized likelihood methods.
\newblock {\em Advances in large margin classifiers} 10(3):61--74.

\bibitem[\protect\citeauthoryear{Ranjan, Le, and
  Hoai}{2018}]{ranjan2018iterative}
Ranjan, V.; Le, H.; and Hoai, M.
\newblock 2018.
\newblock Iterative crowd counting.
\newblock In {\em Proceedings of the European Conference on Computer Vision
  (ECCV)},  270--285.

\bibitem[\protect\citeauthoryear{Ryan \bgroup et al\mbox.\egroup
  }{2009}]{ryan2009crowd}
Ryan, D.; Denman, S.; Fookes, C.; and Sridharan, S.
\newblock 2009.
\newblock Crowd counting using multiple local features.
\newblock In {\em Digital Image Computing: Techniques and Applications, 2009.
  DICTA'09.},  81--88.
\newblock IEEE.

\bibitem[\protect\citeauthoryear{Sam, Surya, and Babu}{2017}]{sam2017switching}
Sam, D.~B.; Surya, S.; and Babu, R.~V.
\newblock 2017.
\newblock Switching convolutional neural network for crowd counting.
\newblock In {\em Proceedings of the IEEE Conference on Computer Vision and
  Pattern Recognition}, volume~1, ~6.

\bibitem[\protect\citeauthoryear{Segu{\'\i}, Pujol, and
  Vitria}{2015}]{segui2015learning}
Segu{\'\i}, S.; Pujol, O.; and Vitria, J.
\newblock 2015.
\newblock Learning to count with deep object features.
\newblock In {\em Proceedings of the IEEE Conference on Computer Vision and
  Pattern Recognition Workshops},  90--96.

\bibitem[\protect\citeauthoryear{Shang, Ai, and Bai}{2016}]{shang2016end}
Shang, C.; Ai, H.; and Bai, B.
\newblock 2016.
\newblock End-to-end crowd counting via joint learning local and global count.
\newblock In {\em Image Processing (ICIP), 2016 IEEE International Conference
  on},  1215--1219.
\newblock IEEE.

\bibitem[\protect\citeauthoryear{Shi \bgroup et al\mbox.\egroup
  }{2018}]{shi2018crowd}
Shi, Z.; Zhang, L.; Liu, Y.; Cao, X.; Ye, Y.; Cheng, M.-M.; and Zheng, G.
\newblock 2018.
\newblock Crowd counting with deep negative correlation learning.
\newblock In {\em Proceedings of the IEEE conference on computer vision and
  pattern recognition},  5382--5390.

\bibitem[\protect\citeauthoryear{Simonyan and
  Zisserman}{2014}]{simonyan2014very}
Simonyan, K., and Zisserman, A.
\newblock 2014.
\newblock Very deep convolutional networks for large-scale image recognition.
\newblock {\em arXiv preprint arXiv:1409.1556}.

\bibitem[\protect\citeauthoryear{Sindagi and Patel}{2017a}]{sindagi2017cnn}
Sindagi, V.~A., and Patel, V.~M.
\newblock 2017a.
\newblock {CNN}-based cascaded multi-task learning of high-level prior and
  density estimation for crowd counting.
\newblock In {\em Advanced Video and Signal Based Surveillance (AVSS), 2017
  14th IEEE International Conference on},  1--6.
\newblock IEEE.

\bibitem[\protect\citeauthoryear{Sindagi and
  Patel}{2017b}]{sindagi2017generating}
Sindagi, V.~A., and Patel, V.~M.
\newblock 2017b.
\newblock Generating high-quality crowd density maps using contextual pyramid
  {CNN}s.
\newblock In {\em Proceedings of the IEEE Conference on Computer Vision and
  Pattern Recognition},  1861–1870.

\bibitem[\protect\citeauthoryear{Srivastava \bgroup et al\mbox.\egroup
  }{2014}]{srivastava2014dropout}
Srivastava, N.; Hinton, G.; Krizhevsky, A.; Sutskever, I.; and Salakhutdinov,
  R.
\newblock 2014.
\newblock Dropout: a simple way to prevent neural networks from overfitting.
\newblock {\em The Journal of Machine Learning Research} 15(1):1929--1958.

\bibitem[\protect\citeauthoryear{Wang and Yung}{2009}]{wang2009crowd}
Wang, L., and Yung, N.~H.
\newblock 2009.
\newblock Crowd counting and segmentation in visual surveillance.
\newblock In {\em Image Processing (ICIP), 2009 16th IEEE International
  Conference on},  2573--2576.
\newblock IEEE.

\bibitem[\protect\citeauthoryear{Wang \bgroup et al\mbox.\egroup
  }{2015}]{wang2015deep}
Wang, C.; Zhang, H.; Yang, L.; Liu, S.; and Cao, X.
\newblock 2015.
\newblock Deep people counting in extremely dense crowds.
\newblock In {\em Proceedings of the 23rd ACM international conference on
  Multimedia},  1299--1302.
\newblock ACM.

\bibitem[\protect\citeauthoryear{Wang \bgroup et al\mbox.\egroup
  }{2019}]{wang2019learning}
Wang, Q.; Gao, J.; Lin, W.; and Yuan, Y.
\newblock 2019.
\newblock Learning from synthetic data for crowd counting in the wild.
\newblock In {\em Proceedings of the IEEE Conference on Computer Vision and
  Pattern Recognition},  8198--8207.

\bibitem[\protect\citeauthoryear{Xie, Noble, and
  Zisserman}{2018}]{xie2018microscopy}
Xie, W.; Noble, J.~A.; and Zisserman, A.
\newblock 2018.
\newblock Microscopy cell counting and detection with fully convolutional
  regression networks.
\newblock {\em Computer methods in biomechanics and biomedical engineering:
  Imaging \& Visualization} 6(3):283--292.

\bibitem[\protect\citeauthoryear{Zhang \bgroup et al\mbox.\egroup
  }{2015}]{zhang2015cross}
Zhang, C.; Li, H.; Wang, X.; and Yang, X.
\newblock 2015.
\newblock Cross-scene crowd counting via deep convolutional neural networks.
\newblock In {\em Computer Vision and Pattern Recognition (CVPR), 2015 IEEE
  Conference on},  833--841.
\newblock IEEE.

\bibitem[\protect\citeauthoryear{Zhang \bgroup et al\mbox.\egroup
  }{2016}]{zhang2016single}
Zhang, Y.; Zhou, D.; Chen, S.; Gao, S.; and Ma, Y.
\newblock 2016.
\newblock Single-image crowd counting via multi-column convolutional neural
  network.
\newblock In {\em Proceedings of the IEEE conference on computer vision and
  pattern recognition},  589--597.

\bibitem[\protect\citeauthoryear{Zhang \bgroup et al\mbox.\egroup
  }{2017}]{zhang2017fcn}
Zhang, S.; Wu, G.; Costeira, J.~P.; and Moura, J.~M.
\newblock 2017.
\newblock {FCN-RLSTM}: Deep spatio-temporal neural networks for vehicle
  counting in city cameras.
\newblock In {\em Computer Vision (ICCV), 2017 IEEE International Conference
  on},  3687--3696.
\newblock IEEE.

\end{thebibliography}
}

\newpage
\appendix
\onecolumn
\noindent{\large \bf Supplementary Material for Crowd Counting with Decomposed Uncertainty}

\section{CNN giga-pixel imagery for the 2017 U.S. Presidential inauguration}

Earlier in the paper, we raised a question of how much we can trust predictions of a model, especially when we do not have labels or ground truth to verify the accuracy of the predictions. Now with uncertainty estimates at hand, we can present crowd counting predictions on new real world data. In the supplementary material, we show our results on CNN (Cable News Network) giga-pixel images which contains  ultra high resolution (64,000 $\times$ 64,000 pixels) crowd images.

\begin{figure}[ht]
    \centering
    \includegraphics[width=0.82\linewidth]{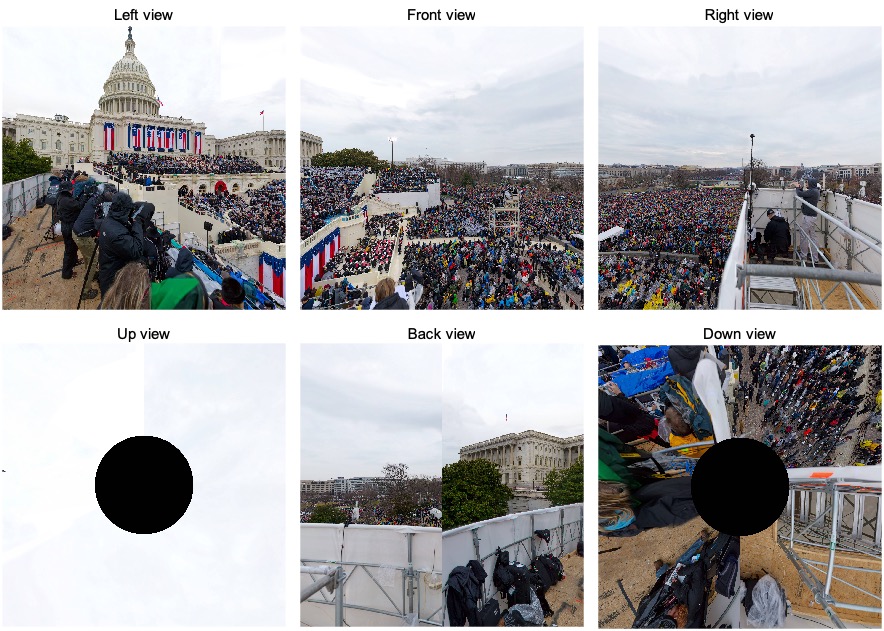}
    \caption{Different viewing directions of the CNN giga-pixel images
    }
     \label{fig:gigapixel}
\end{figure}

 CNN released the giga-pixel images of the 2017 U.S. Presidential inauguration \cite{CNNgigapixel}. The CNN giga-pixel images consist of the photos taken from 6 different viewing directions: left, front, right, up, back and down views as shown in Figure \ref{fig:gigapixel}. Each viewing direction contains 15,625 (125 $\times$ 125) patches of 500 $\times$ 500 pixel images, which means that each viewing direction contains a total of approximately $3.9 \times 10^9$ pixels (3.9 giga-pixels). Additionally, the zoomed out versions (with 7 different zoom levels) of the images are also included. However, in fact they are stitched images of the original high resolution images (not additional photos). 

To the best of our knowledge, this is the largest high resolution crowd image dataset from a single event. We present crowd counting prediction results on these giga-pixel images using our proposed method, DUBNet. Clearly, the dataset does not contain the ground truth labels or aggregated counts. Hence, merely performing a point estimation of counts (or density maps) on these images is not sufficient. In the evaluations on the benchmark datasets we presented in the paper, DUBNet not only demonstrates the state of art performance in terms of counting accuracy but also presents an insightful uncertainty measures for its prediction. With this uncertainty estimates at hand, we can perform predictions and measure how confident we are in our predictions. 

We observe that the majority of the crowd are captured in the three views: left, front and fright views. Hence, we present the results from these three views. We use a network that was trained on the UCF-QNRF dataset for prediction. 

\begin{figure*}[t]
    \centering
    \includegraphics[width=0.90\linewidth]{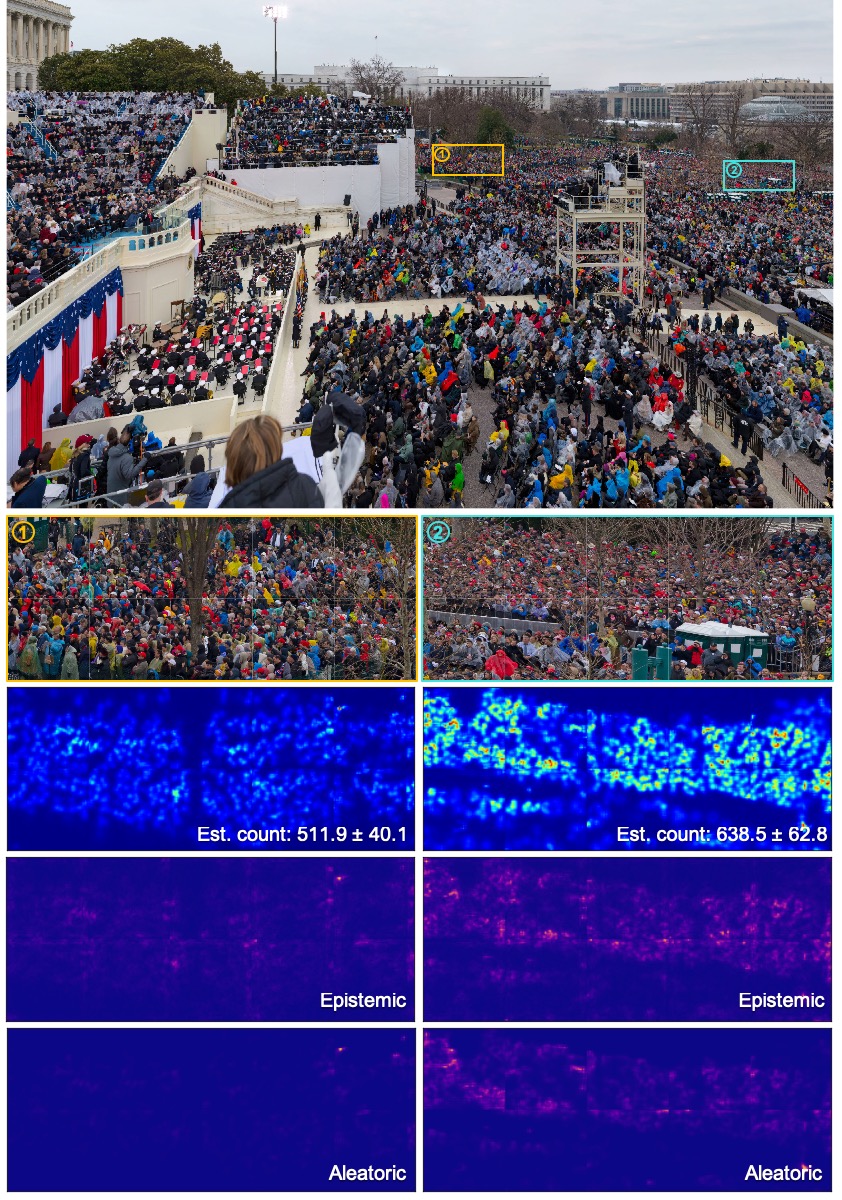}
    \caption{The front view of the CNN giga-pixel images and prediction examples
    }
     \label{fig:trump_f}
\end{figure*}

Note that while the method attempts to predict the count of the people present in the images, this is ``not'' about estimating a total number of attendees at the inauguration. We acknowledge that even if we could count accurately every person present in the image, that estimated number of people would be still smaller than the actual number of people who attended the inauguration since there are parts of the event that the giga-pixel images are not able to capture. For example, in Figure \ref{fig:trump_f}, a significant portion of the crowd is blocked from the view by the white temporary structure on which the camera operators are standing. We do not attempt any inference on what is not seen in the images.

\begin{figure*}[t]
    \centering
    \includegraphics[width=0.85\linewidth]{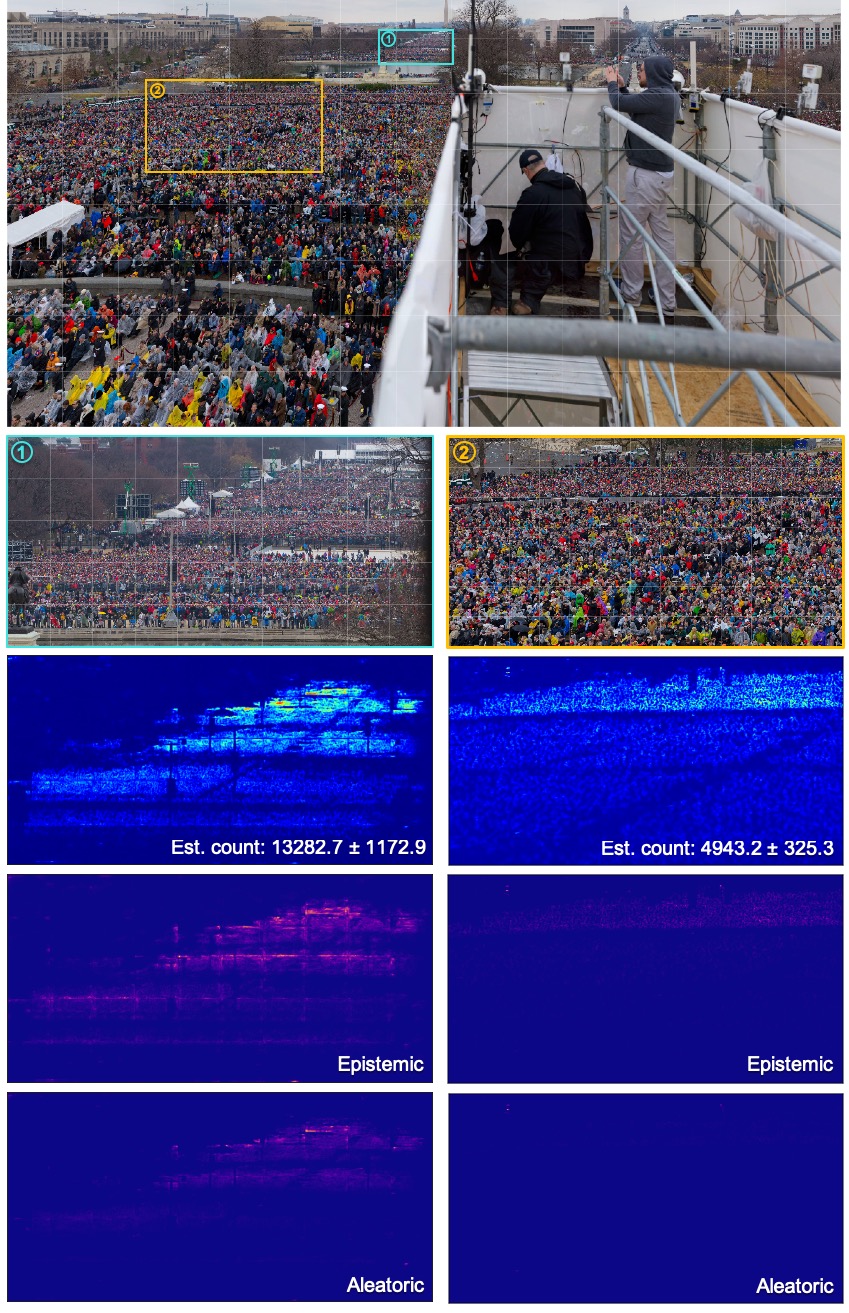}
    \caption{The right view of the CNN giga-pixel images and prediction examples
    }
     \label{fig:trump_r}
\end{figure*}

\begin{figure*}[t]
    \centering
    \includegraphics[width=0.90\linewidth]{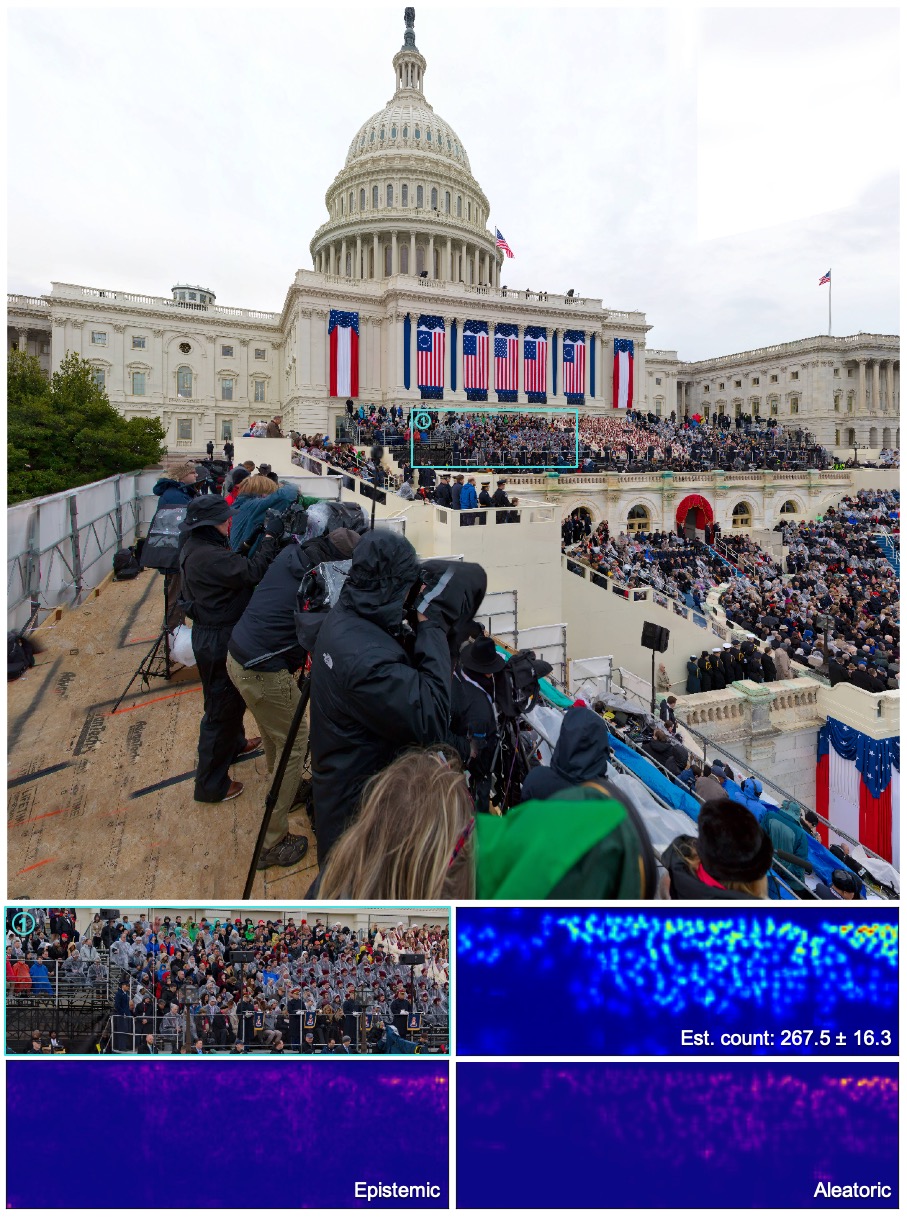}
    \caption{The left view of the CNN giga-pixel images and prediction examples
    }
     \label{fig:trump_l}
\end{figure*}

\clearpage

While one could simply apply DUBNet (or any other neural network based crowd counting method) directly to the giga-pixel images with each single direction image collage being a one large image input, it would not be a good practice. That is because it is highly likely to encounter a memory issue due to the large size and also because the trained network has not seen a person in such a larger scale (for example, consider the pixel sizes of cameramen on the bottom left region of Figure \ref{fig:trump_l}). One can pass each image with a fixed zoom level as input to the network. The results using a different zoom level (but fixed throughout prediction) are shown in Table \ref{gigapixel_pred_naive}. The estimated counts are provided with the 90\% confidence interval. Note that since the ground-truth density or counts do not exist, we used the calibration method validated on UCF-QNRF test dataset.\footnote{In order to compute an accurate estimate of confidence intervals, perhaps this is the minimal portion that a practitioner can provide labeled data if available. However, for the sake of completeness of our presentation without any ground truth labels provided, we proceed this way.}


\begin{table}[ht]
\begin{center}
\begin{tabular}{|l|c|c|c|c|}
\hline
&  Left view &  Front view & Right view & Total\\
Zoom & Est. count  &  Est. count   & Est. count  & \\
\hline\hline
Level 3  & 1121.5 $\pm$ 219.8		& 7390.6	$\pm$ 837.4   &	8894.6	$\pm$ 1702.5 & 17906.7 $\pm$ 2759.6\\
Level 4  & 2515.1 $\pm$ 204.9		& 9702.4	$\pm$ 944.0   &	14419.1 $\pm$ 2637.4 & 26636.6 $\pm$ 3786.3\\
Level 5  & 5358.9 $\pm$ 385.3		& 14444.2	$\pm$ 1220.8  &	25602.2 $\pm$ 4164.1 & 45405.3 $\pm$ 5770.2\\
Level 6  & 18985.1 $\pm$ 967.2		& 34130.1	$\pm$ 2914.3  &	48807.1 $\pm$ 9328.2 & 101922.3 $\pm$ 13209.7\\
Level 7  & 68482.5 $\pm$ 2636.8	    & 69880.6	$\pm$ 5836.7  &	61582.8 $\pm$ 11294.3 & 199945.9 $\pm$ 19767.8\\
\hline
\end{tabular}
\end{center}
\caption{Zoom levels and predictions}
\label{gigapixel_pred_naive}
\end{table}

We observe that the estimated counts monotonically increase with the zoom level, which is expected, since the higher the zoom level (higher resolution) is the larger the number of pixels is. Furthermore, since output pixel values were trained to be non-negative (since the prediction output is a density map), it is highly likely that as zoom level increases, the counts will lead to a higher positive bias if we do not use any masking to filter areas that are not regions of interest (ROI). Consider Figure \ref{fig:trump_l} for example. The majority of the images patches in this viewing direction may not include any person at all. ROI does not have to be a precise topology over crowd regions but rather rectangular partition that include at least a few people (not just  body parts) would suffice. 
Clearly, the total estimated count for zoom level 7 is an overestimation --- for example, one can easily point out that predicting more than 68,000 people in Figure \ref{fig:trump_l} seems too high even at a first glance. Also, the estimated crowd counts for zoom level 3 seem too low. Then, does the right zoom level exist somewhere between 3 and 7? The answer is no. Table \ref{gigapixel_pred_naive} shows that there is no single zoom level which works best for all viewing directions.

Therefore, we adaptively choose a zoom level given a region of an image. To be consistent with the training data (UCF-QNRF), we require each image patch size to be large enough (i.e. zoomed out enough) to have at least 20 people for a given region. Of course, we do not have the ground truth counts. Hence we use predicted counts to adjust the zoom level. If the threshold is not satisfied according to the predicted count, we zoom out, i.e. zoom level decreases and the image patch containing the region of interest grows. If the threshold is satisfied, then we use the lowest predictive variance to decide which zoom level for the region is optimal. By iteratively following the procedure, we form partitions over an image and report predicted counts (predicted density map) per each partition. The sample results are shown in Figures \ref{fig:trump_r}, \ref{fig:trump_f}, and \ref{fig:trump_l}. Each colored square in an image collage represents a sample partition chosen for particular regions. Their corresponding predictive density maps and uncertainty estimates are shown in each subplot.


\begin{table}[ht]
\begin{center}
\begin{tabular}{|c|c|c|c|}
\hline
  Left view &  Front view & Right view &  Total\\
Est. count & Est. count & Est. count &  \\
\hline\hline
 1202.7 $\pm$ 91.8 & 12408.9 $\pm$ 1178.3 & 34714.5 $\pm$ 6207.0 & 48326.1 $\pm$ 7477.1\\
\hline
\end{tabular}
\end{center}
\caption{Predictions on CNN giga-pixel images}
\label{gigapixel_pred_final}
\end{table}

We acknowledge the prediction on first sample partition (1) in Figure \ref{fig:trump_r} is very challenging and perhaps the predicted count is an underestimation. However, we used zoom level 7 (the highest resolution) for this partition and can't improve further. However, most of other partitions demonstrate plausible density outputs and count estimates. Table \ref{gigapixel_pred_final} report the aggregated count results over the partitions for each viewing direction with the corresponding confidence intervals. Again, in this analysis, we only report on what our method predicted based on what it sees and do not make any inference on what is not seen on the images.

\section{Discussions on Bayesian neural network}
In this section we discuss the Bayesian neural network methods, and the justification for the use of bootstrap ensemble to approximate the posterior in this work.
Let $\mathcal{D} = \{ (\bm{x}_i, \bm{y}_i)\}^N_{i=1}$ be a collection of realizations of i.i.d random variables, where $\bm{x}_i$ is an image, $\bm{y}_i$ is a corresponding density map, and $N$ denotes the sample size. In Bayesian neural network framework, rather than thinking of the weights of the network as fixed parameters to be optimized over, it treats them as random variables, and so we place a prior distribution $p(\theta)$ over the weights of the network $\theta \in \theta$. This results in the posterior distribution
\begin{equation*}
    p(\theta | \mathcal{D}) = \frac{p(\mathcal{D} | \theta)p(\theta)}{p(\mathcal{D})} = \frac{\left(\prod^N_{i=1} p(y_i | x_i, \theta)\right) p(\theta)}{p(\mathcal{D})}.
\end{equation*}
While this formalization is simple, the learning is often challenging because calculating the posterior $p(\theta | \mathcal{D})$ requires an integration with respect to the entire parameter space $\Theta$ for which a closed form often does not exist. \cite{mackay1992practical} proposed a Laplace approximation of the posterior.
\cite{neal1993bayesian} introduced the Hamiltonian Monte Carlo, a Markov Chain Monte Carlo (MCMC) sampling approach using Hamiltonian dynamics, to learn Bayesian neural networks. This yields a principled set of posterior samples without direct calculation of the posterior but it is computationally prohibitive. 
Another Bayesian method is variational inference \cite{blundell2015weight,graves2011practical,louizos2016structured,louizos2017multiplicative} which approximates the posterior distribution by a tractable variational distribution $q_\eta(\theta)$ indexed by a variational parameter $\eta$. The optimal variational distribution is the closest distribution to the posterior among the pre-determined family $Q = \{q_\eta(\theta)\}$. The closeness is often measured by the Kullback-Leibler (KL) divergence between $q_\eta(\theta)$ and $p(\theta | \mathcal{D})$. While these Bayesian neural networks are the state of art at estimating predictive uncertainty, these require significant modifications to the training procedure and are computationally expensive compared to standard (non-Bayesian) neural networks 

\cite{gal2016dropout} proposed using Monte Carlo dropout to estimate predictive uncertainty by using dropout at test time. There has been work on approximate Bayesian interpretation of dropout \cite{gal2016dropout,kingma2015variational,maeda2014bayesian}. Specifically, \cite{gal2016dropout} showed that Monte Carlo dropout is equivalent to a variational approximation in a Bayesian neural network. With this justification, they proposed a method to estimate predictive uncertainty through variational distribution. Monte Carlo dropout is relatively simple to implement leading to its popularity in practice. Interestingly, dropout may also be interpreted as ensemble model combination \cite{srivastava2014dropout} where the predictions are averaged over an ensemble of neural networks. The ensemble interpretation seems more plausible particularly in the scenario where the dropout rates are not tuned based on the training data, since any sensible approximation to the true Bayesian posterior distribution has to depend on the training data. This interpretation motivates the investigation of ensembles as an alternative solution for estimating predictive uncertainty. Despite the simplicity of dropout implementation, we were not able to produce satisfying confidence interval for our crowd counting problem. Hence we consider a simple non-parametric bootstrap of functions which we discuss in Section \ref{bootstrap}.

\section{Ground truth generation}\label{sec:gt_generation}
We generate the ground truth density maps by blurring the head annotations provided by the data. This blurring is done by applying a Gaussian kernel (which normalize to 1) to each of the heads in a given image. We use geometry-adaptive kernels \cite{zhang2016single} to vary the spread parameters of Gaussian depending on local crowd density. The geometry-adaptive kernel is given by:
\begin{equation*}
    F(z) = \sum_{j=1}^J \delta(z - z_j) \times G_{\sigma_j} (z), \text{ with } \sigma_j = \beta \bar{d}_j
\end{equation*}
For each targeted object $z_j$ in the ground truth $\delta$, we use $\bar{d}_j$ to indicate the average distance of $k$ nearest neighbors. To generate the density map, we convolve $\delta(z - z_j)$ with a Gaussian kernel with parameter $\sigma_j$ (standard deviation),
where $z$ is the position of pixel in the image.

\begin{figure}
    \centering
    \includegraphics[width=\linewidth]{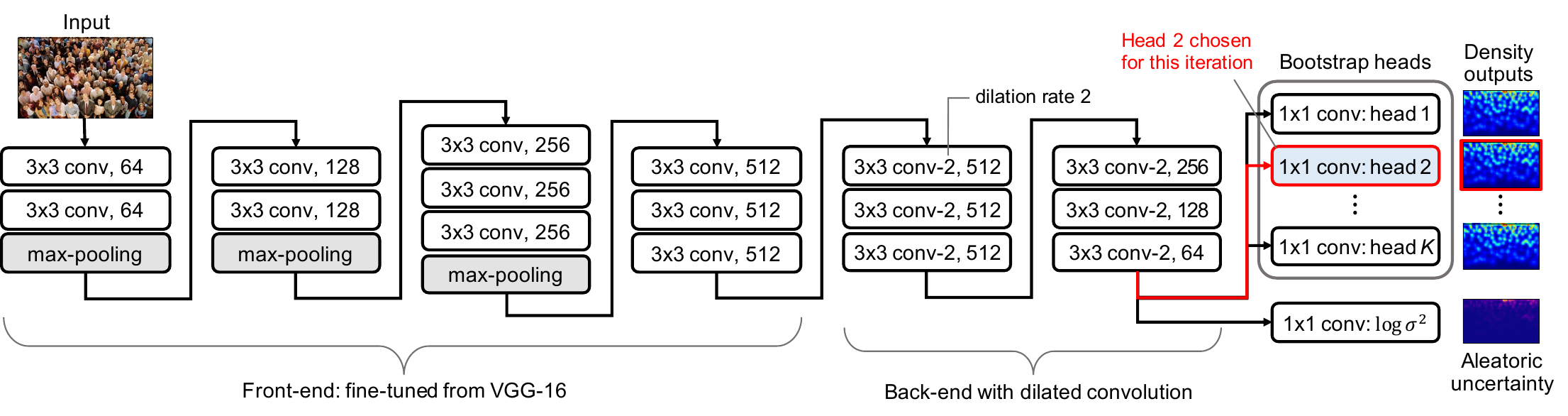}
    \caption{Network architecture of our proposed method, DUBNet. All convolutional layers use
SAME padding to maintain the size of the output the same as the input size. Max-pooling layers are applied with 2$\times$2 window with stride size 2. The back-end layers use dilated kernels with rate 2. The output layer branches out to $K$ bootstrap heads and an extra log-variance output. 
}
    \label{fig:CSRNet_architecture}
\end{figure}

\section{Comparison on variability}
Note that we do not have the ground truth uncertainty to evaluate the predictive uncertainty other than testing whether predictive uncertainty satisfies the definition of confidence interval discussed in Section \ref{section:uncertainty_calibration}.
While we recalibrate the estimated uncertainty, we perform a sanity check on the amount of variability of our proposed method before recalibration is applied.
We compare our proposed framework with a full bootstrap ensemble, i.e. an ensemble of $K$ independent neural networks. Although hypothetically $K$ bootstrap ensembles can lead to $K$ identical models in the worst cases, due to the nature of highly non-linear objective in neural network parameter optimization along with random initialization, we should not be worried about this degenerate case. 
Note that the architectural setting of DUBNet has a minimal bootstrapping with each output head only branching out at the end of the network architecture, which achieves computational gains but could potentially limit this variability. Hence, we compare our method with a full bootstrap ensemble model which contains the $K$ full-size independent neural networks with  each neural network being trained independently. We compute the average of estimated predictive variance on ShanghaiTech Part A and Part B test datasets. In Table \ref{variance_test}, we report the predictive variance which is the sum of epistemic and aleatoric uncertainties before recalibration is applied. we observe that DUBNet shows slightly lower variance than the full bootstrap model, which is expected since the amount of the shared portion of the architecture is much higher for DUBNet. However, surprisingly the difference in variance is not much given the contrasting network sizes between the full bootstrap model and DUBNet. Most importantly, with recalibration procedure at hand, we can correct (amplify or shrink) the predictive variance to a suitable amount. Hence, the post-calibrated uncertainty for these two models are almost identical given the same validation dataset and the test dataset.
\begin{table}[ht]
\begin{center}
\begin{tabular}{|l|c|c|}
\hline
& \multicolumn{2}{c|}{Variance}\\
Method & Part A & Part B \\
\hline\hline
Full-bootstrap ResNet frontend + Dilated backend & 47.6 & 1.17 \\
DUBNet (Ours) & 45.8 & 1.02  \\
\hline
\end{tabular}
\end{center}
\caption{Comparison on average predictive variance }
\label{variance_test}
\end{table}

\section{Note on Inference Runtime}
We tested the inference runtime of DUBNet compared to the DUBNet without the DUB extension (i.e., single branch 1-1 connected layer) to see how much computational increase the proposed bootstrap extension causes. The first set of tests were performed on a CPU with 2.3GHz quad-core Intel Core i5 (8GB RAM). Nvidia P100 GPU was used for the second set of the tests. Each test was performed on ShanghaiTech Part A test dataset. 
The test showed that the average additional cost is very minimal, with 2\% increase on CPU and 0.5\% increase on GPU (almost negligible). This makes sense since the output heads only cover the last layer. Also, note that for training since we sample one output node per epoch and update the weights accordingly, there is no additional computation cost per epoch. Hence, this supports our claim that the proposed method is a very efficient way of producing uncertainty estimate.

\end{document}